\documentclass{isprs} 
\usepackage{subfigure}
\usepackage{setspace}
\usepackage{geometry} 
\usepackage{epstopdf}
\usepackage[labelsep=period]{caption}  
\usepackage[british]{babel} 
\usepackage[hang]{footmisc}
\usepackage[noend,ruled,vlined]{algorithm2e}
\usepackage{amsmath}
\usepackage{amsfonts}
\usepackage{multirow}
\usepackage{natbib}
\usepackage{booktabs}
\usepackage{stfloats}
\usepackage{wasysym}

\begin{document}

\title{WEAKLY  SUPERVISED PSEUDO-LABEL ASSISTED LEARNING FOR ALS  POINT CLOUD SEMANTIC SEGMENTATION
}
\version{}

\author{Puzuo Wang$^{1}$, Wei Yao$^{1}$\thanks{Corresponding author.}}

\address{$^{1}$ Department of Land Surveying and Geo-Informatics, The Hong Kong Polytechnic University, Hung Hom, Hong Kong\\
- puzuo.wang@connect.polyu.hk -wei.hn.yao@polyu.edu.hk}


\commission{II, }{3} 
\workinggroup{II/3} 
\icwg{}   

\abstract{
Competitive point cloud semantic segmentation results usually rely on a large amount of labeled data. However, data annotation is a time-consuming and labor-intensive task, particularly for three-dimensional point cloud data. Thus, obtaining accurate results with limited ground truth as training data is considerably important. As a simple and effective method, pseudo labels can use information from unlabeled data for training neural networks. In this study, we propose a pseudo-label-assisted point cloud segmentation method with very few sparsely sampled labels that are normally randomly selected for each class. An adaptive thresholding strategy was proposed to generate a pseudo-label based on the prediction probability. Pseudo-label learning is an iterative process, and pseudo labels were updated solely on ground-truth weak labels as the model converged to improve the training efficiency. Experiments using the ISPRS 3D sematic labeling benchmark dataset indicated that our proposed method achieved an equally competitive result compared to that using a full supervision scheme with only up to 2$\permil$ of labeled points from the original training set, with an overall accuracy of 83.7$\%$ and an average F1 score of 70.2$\%$.
}


\maketitle


\section{INTRODUCTION}\label{INTRODUCTION}
 
\sloppy

The recent development of convolutional neural networks (CNNs) has been followed by the progress of point cloud semantic segmentation tasks, and there are constantly new methods that achieve state-of-the-art results. Most of these methods focus on designing new network structures or convolution kernels based on the characteristics of point cloud data~\citep{hu2020randla,thomas2019kpconv,huang2020deep,steinsiek2017semantische}, whereas few focus on data labeling. In experiments, competitive results rely on a large amount of data annotations, which continuously require labeling operators with expert knowledge. The labeling of point cloud data is particularly difficult, and the operator usually needs to confirm the category of a point from multiple perspectives. Although labeling work is difficult and time-consuming, obtaining raw point cloud data has become easier. With the advancement of light detection and ranging sensor technology and the diversification of data acquisition platforms, obtaining massive unlabeled point cloud data is no longer a problem. Thus, extracting information that helps the task of semantic segmentation from unlabeled data is an essential issue. By using a small amount of annotated data to achieve classification results comparable to full supervision, the workload of data annotation would be considerably reduced and the efficiency in related applications would be improved. 

\begin{figure}[ht!]
\begin{center}
       \subfigure[]{
                \begin{minipage}{.45\linewidth}
                \centering
                \includegraphics[width=1.0\columnwidth]{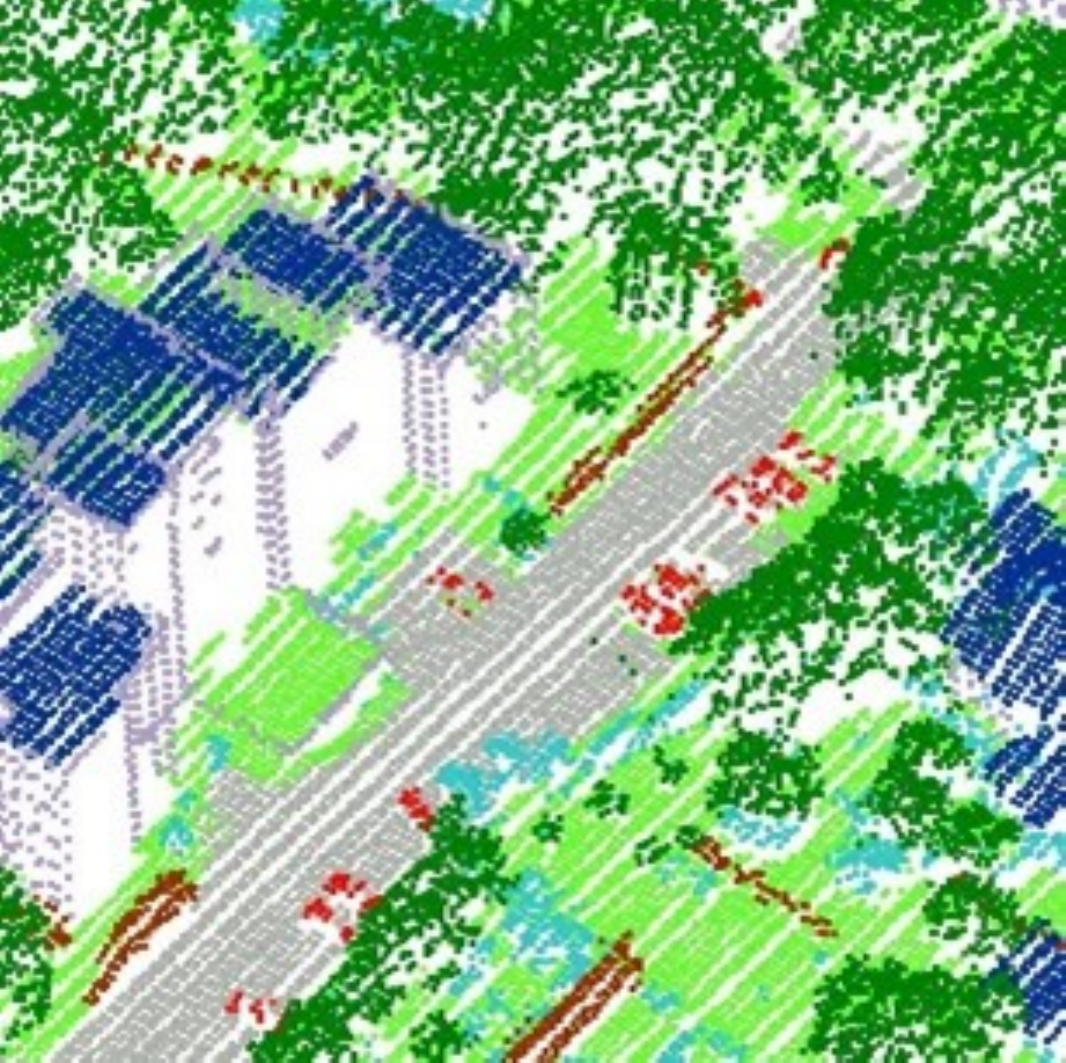}
                \end{minipage}
                }
      \subfigure[]{
                \begin{minipage}{.45\linewidth}
                \centering
                \includegraphics[width=1.0\columnwidth]{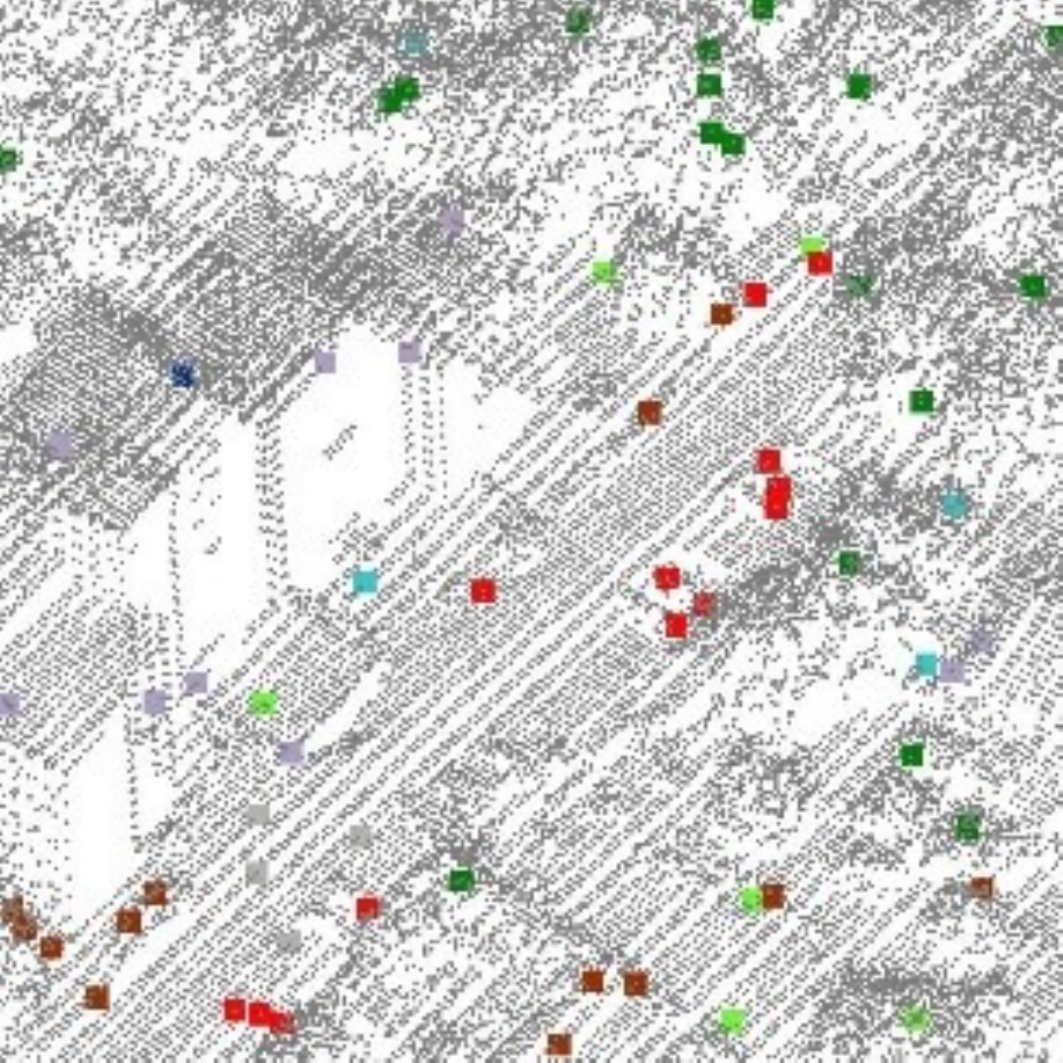}
                \end{minipage}
                }               
	\caption{Comparison of (a) fully-labeled points and (b) weak-label. (b) Situation in which only 100 points per class ($2\permil$) are labeled. The size of labeled points in (b) is enlarged for better visualization.}
\label{fig:figure1}
\end{center}
\end{figure}

Semi- and weakly supervised learning are commonly used to address situations in which the number of labels is scarce. There are some comprehensive reviews on these two types of supervised learning methods~\citep{chapelle2009semi,zhou2018brief,zhu2005semi}. Choosing a suitable weak label annotation strategy is key to balancing the labeling workload and experimental results. In image processing, weak labels are represented as a few labeled images~\citep{dong2018few}, a few labeled pixels~\citep{bearman2016s}, and a few bounding boxes or categories that appear on the images~\citep{kolesnikov2016seed,papandreou2015weakly}. Until now, there have been very few works that used weakly supervised methods to process point cloud data.~\citet{wei2020multi} applied a point class activation map to classify point clouds using only cloud-level labels. However, we believe that it is not practical for point cloud data, particularly in outdoor scenes that cover a wide region, because the complete data must be divided into many small blocks and categories contained in each block must be specified. In comparison, assigning labels to a few points is a more direct approach.~\citet{polewski2015active} used an active learning method to detect standing dead trees from airborne laser scanning (ALS) point cloud data combined with infrared images.~\citet{lin2020active} proposed an active and incremental learning strategy for ALS semantic segmentation, and manual annotation was iteratively added for training. Nonetheless, the setting of a weak label is used to annotate all points of tiles, and manual intervention is required during training. Through theoretical analysis and experimental results,~\citet{xu2020weakly} found that when the number of labeled points remains constant, the spatially sparse annotations achieve better results than those samples gathering in certain object instances, and this weak-label setting was adopted in this work. An illustration of the sparse weak-label situation is shown in Fig.~\ref{fig:figure1}. A weakly supervised semantic point cloud segmentation framework was proposed by~\citet{xu2020weakly}, and an approximate result of fully supervised learning was obtained using 10$\%$ labels. However, the characteristic of this weak label can be understood as being spatially continuous at a lower resolution, and the workload of labeling is still large.~\citet{guinard2017weakly} utilized the point cloud segmentation method to improve the classification accuracy with very few labels, but the result largely depends on the segmentation accuracy and focuses on classifying the point cloud of the area where the initial weak label is located, which is transductive learning. 

Pseudo-labels, assigning annotations to unlabeled data based on the predictions of the current model, can enable the use of unlabeled data in parameter updates. Through the pseudo-label method, the classification model can calculate the loss function and backpropagation from the dataset with more labels, thereby improving the accuracy.~\citet{iscen2019label} and~\citet{lee2013pseudo} applied pseudo-labels in image classification, and~\citet{zou2020pseudoseg} designed pseudo-labeling and data augmentation to improve the performance of image semantic segmentation.~\citet{NEURIPS2019_1cd138d0} mixed several proven semi-supervision strategies and proposed a holistic framework. Only a few studies have utilized pseudo-labels in point cloud processing.~\citet{yao2020pseudo} introduced a pseudo-labeling method into point cloud semantic segmentation. However, similar to~\citet{guinard2017weakly}, the framework is also transductive learning , and the performance of the model has not been verified on untrained test data.

To cooperate with pseudo-labels, it is essential to propose an effective semantic segmentation network structure. Owing to the irregular distribution of point clouds, the network structure is not as uniform as that of two-dimensional (2D) CNNs. Early processing methods project point clouds onto images and directly use mature 2D CNNs to train the model~\citep{boulch2018snapnet}. Another branch of methods voxelizes point clouds and proposes three-dimensional (3D) CNNs to process voxel data ~\citep{tchapmi2017segcloud}. The disadvantage of these two methods is that as point clouds need to be converted into regularized data, geometric information may be lost. PointNet and PointNet++~\citep{qi2017pointnet,qi2017pointnet++} are pioneers in the development of shared multilayer perceptrons (MLPs) to directly analyze point clouds. However, the spatial distribution relationship of the point cloud requires well-designed MLPs, which complicates network structures. Compared to pointwise MLP networks, graph convolution networks construct a graph through relative spatial positions between points for feature extraction and fusion~\citep{wang2019dynamic}. Point convolution networks are similar to graph-based methods, in which point kernels are designed to learn local geometric information~\citep{thomas2019kpconv}.

In this study, we explored how to obtain reliable semantic segmentation results with very few labels. A pseudo-label-assisted point cloud semantic segmentation framework with extremely few annotations is proposed. We used KPConv~\citep{thomas2019kpconv}, a point convolution network, as our encoder network. Our weak label was defined as sparsely labeled points randomly distributed in space. An initial model was trained using the selected weak labels. Pseudo-labels were then generated by the trained model. Considering that the model obtained from weak label training is underfitting  to the entire data space, an adaptive threshold is proposed to balance the number and accuracy of pseudo-labels. The training procedure that combines the ground-truth labels, referred to weak labels, and pseudo-labels was iteratively performed. To accelerate the training progress and reduce the influence of pseudo labels containing errors on the model, we updated the pseudo labels when the model converged on the ground-truth labels. Experiments on the ALS dataset showed the effectiveness of the method. 

\section{METHODOLOGY}\label{sec:METHODOLOGY}

\begin{figure}[b!]
\captionsetup{justification=centering}
\begin{center}
		\includegraphics[width=0.7\columnwidth]{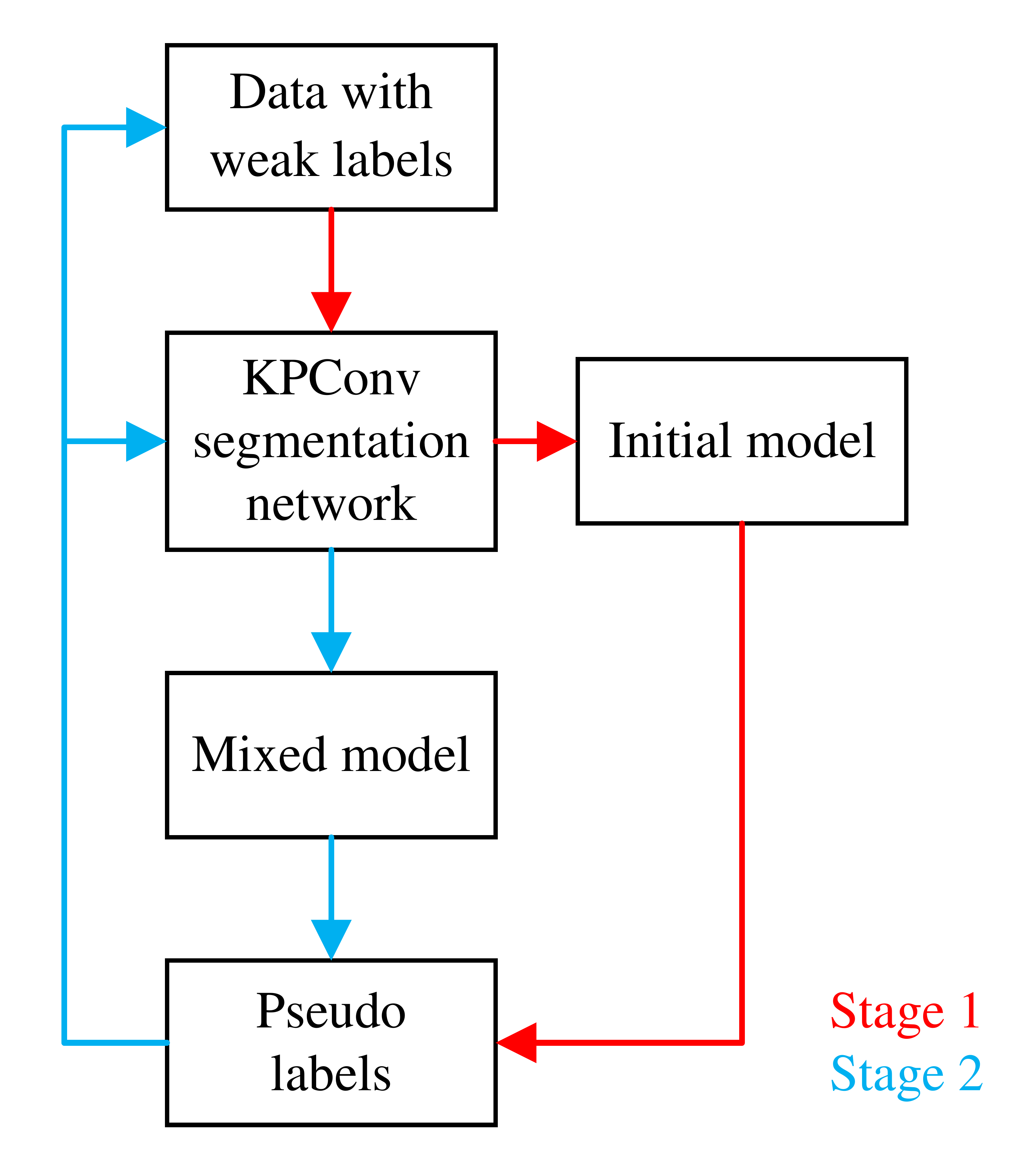}
	\caption{Architecture of the proposed method.}
\label{fig:figure2}
\end{center}
\end{figure}

An overview of the proposed method is presented in Fig.~\ref{fig:figure2}. There are two stages in the framework: red lines represent incomplete training and blue lines represent pseudo-label-assisted training. In stage 1, the initial model is trained using sparse labeled points, based on which pseudo labels are generated. In stage 2, the training progress is continued with a combination of the initial ground truth and pseudo labels. Pseudo labels are iteratively updated once the mixed trained model converges on the labels. The following sections include the introduction of the semantic segmentation network, the generation and update of pseudo labels, and the entire training process.

\subsection{Point Cloud Semantic Segmentation Network}\label{sec:Point Cloud Semantic Segmentation Network}

This section briefly reviews the architecture of the adopted network. In this work, KPConv was used as our encoder network because of its state-of-the-art experimental results on several public datasets. KPConv uses the idea of convolution kernels in image convolutions and can extend kernel points to be deformable to adapt the local features of point clouds. We chose the rigid point convolution kernel and used the same network architecture as KPConv for the semantic segmentation task. The encoder network contained five convolutional layers, and the ResNet-like structure was embedded. Skip links were used in the decoder network, and features were passed by the nearest sampling.

\subsection{Incomplete Supervision}\label{sec:Incomplete Supervision}

In this study, weak labels are defined as limited and sparse ground-truth labels distributed across the space, which is referred to as incomplete supervision. In the form of manual annotation , we randomly selected several points according to the category and assigned the ground-truth labels. Given that the input points $P\in\mathbb{R}^{N\times D}$ consist of N points with D dimensional features, M(M$\ll$N) points are assigned labels, denoted as $P_w\in\mathbb{R}^{M\times D}$ and the corresponding labels $L_w\in\mathbb{R}^{M}$. 

The training process of incomplete supervision is similar to that of full supervision, and the only difference lies in the design of the loss function. Because only a few points have label information, we calculated the loss of these points and performed backpropagation. We chose the softmax cross-entropy loss function on the labeled points, denoted as:
\begin{equation}\label{equ:1}
	l_{true} =  - \frac{1}{M}\sum\limits_i {{{\hat y}_{ic}}} \frac{{\exp({y_{ic}})}}{{\sum\nolimits_c {\exp({y_{ic}})} }},\
\end{equation}
where $y_{ic}$ and ${\hat y}_{ic}$ are the prediction and label of point $p_{i}$, respectively.

\subsection{Pseudo-Label assisted Learning}\label{sec:Pseudo-Label Assisted Learning}

In incomplete learning, model parameters are trained by partial points, and the accuracy of prediction is significantly lower than that of fully supervised learning. This is because limited labeled points cannot describe the overall distribution of the data. Thus, using more training data can achieve predictions with higher accuracy. Pseudo-labels can effectively exploit unlabeled data as a semi-supervised learning method. The pseudo-label-assisted learning method predicts the unlabeled data through the model parameters obtained by the previous training and regards those labels as ground truth in back propagation calculation~\citep{lee2013pseudo}. By increasing the number of pseudo-labels, we intended to simulate the data distribution of the entire scene. We propose a framework for pseudo-label assisted point cloud semantic segmentation, and the details of the generation and update of pseudo-labels are introduced in the following two sections.

\subsubsection{Generation of Pseudo-labels}\label{sec:Generation of Pseudo-labels}

We used the predictions at the unlabeled data of the training sets to generate pseudo-labels, and the category of each point was determined as the one with the maximum posterior probability:
\begin{equation}\label{equ:2}
	L_{pred} = \arg\max {p_{ic}},\
\end{equation}
Because they are pseudo labels, there are inevitably some incorrect predictions. This is because the parameters of the trained model are only obtained from limited labeled data, which is underfitting to the entire data. Therefore, it is necessary to perform label selection to obtain suitable pseudo-labels. Generally, a prediction with a higher posterior probability is more likely to be correct. Thus, a commonly used method is to set a strict threshold and select labels with predicted probabilities above it. Although a high threshold will result in more accurate pseudo-labels, it will lead to a decrease in the number of labeled points. However,~\citet{yao2020pseudo} found that the experimental results are not sensitive to the threshold setting. Unlike using a fixed value, we developed an adaptive threshold-setting method. After attaining the probability of points toward each predicted category, we chose their average value as the threshold:
\begin{equation}\label{equ:3}
	N_{pseudo} = N({p_i} > \frac{1}{n}\sum\limits_i {{p_i})},\
\end{equation}
where ${p_i} = \max({p_{ic}})$. N and n are the index and the number of predictions of the training data, respectively. Predictions with a probability above the threshold are considered reliable and are selected as pseudo labels. 

\subsubsection{Update of Pseudo Labels}\label{sec:Update of Pseudo Labels}

We aimed to enhance the performance of the classifier through iterative training and identify essential pseudo-labels, thus the previous ones were discarded with the update. The initial pseudo-labels were generated by the model obtained from incomplete supervision, and subsequent updates were generated by pseudo-label-assisted learning. The real labels from the ground truth are essential. Thus, to maintain their influence on the model training, we used the loss function in~\citet{lee2013pseudo}, which calculates the loss of ground truth and pseudo-labels separately. A cross-entropy loss was still utilized for pseudo-labels, denoted as: 
\begin{equation}\label{equ:4}
        l_{pseudo} =  - \frac{1}{{M'}}\sum\limits_i {{{\hat y'}_{ic}}} \frac{{exp({y_{ic}})}}{{\sum\nolimits_c {exp({y_{ic}})} }},\
\end{equation}
where ${\hat y'}_{ic}$ and $M'$ are pseudo-labels and their numbers, respectively. Thus, the total loss in pseudo-label-assisted learning is defined as:
\begin{equation}\label{equ:5}
        L_p = l_{true} +  \alpha  \cdot l_{pseudo},\
\end{equation}
where $\alpha$ is the weight coefficient. Although the number of ground truth and pseudo-labels is quite different, both $l_{true}$ and $ l_{pseudo}$ are averaged loss values over all the points, thus was empirically set to 1. The pseudo-labels were updated when the model fitted the current labels.

Unlike the setting in~\citet{yao2020pseudo}, the model that converges on the ground truth is used as the condition for updating pseudo labels. We believe that there are some errors in pseudo labels, particularly when the scene is complex, thus completely fitting pseudo-labels may result in error transmission. In addition, because $l_{true}$ is retained in the entire training process, convergence on the ground truth is sufficient to guarantee the performance of the model. Another advantage is that it can increase the frequency of pseudo-label updates and improve training efficiency. A comparison is presented in the Experimental section. Convergence was determined by the minimum training accuracy in one epoch. If the value was above 99$\%$, the pseudo labels were updated at the end of the epoch. 

\begin{algorithm}[b]
	\caption{Pseudo-label assisted Point Cloud Semantic Segmentation}
	\label{alg:algorithm1}
	\KwIn{Point clouds ${P\in\mathbb{R}^{N\times D}}$, Labels ${L\in\mathbb{Z}^{M}}$}
	\KwOut{Predictions ${Y\in\mathbb{Z}^{N}}$}  
	
	Initialize ${\Theta}$ randomly; \\
	\# Incomplete supervision\\
	\For{$epoch\gets1$ \KwTo $100$}{
	    Train One Epoch:\\
	    $\Theta_T = \Theta_T - \eta \nabla l_{true}(L,{\rm{ }}Y|\Theta_T)$;\\
	    \# $\Theta$ is learned parameters}
    \# Pseudo-label assisted learning\\
    $\Theta_P=\Theta_T$;\\
    \For{$epoch\gets1$ \KwTo $100$}{
        Pseudo-labels $L' = Y|(P,\Theta_P) \cap Selection$;\\
        \Repeat{Convergence(L)}{
        Train One Epoch:\\
        $\Theta_P =\Theta_P - \eta (\nabla l_{true}(L, Y|\Theta_P)$ \\ 
        \qquad \qquad \qquad \ \ + $\nabla l_{pseudo}(L', Y|\Theta_P))$;}
        }
\end{algorithm}

\subsection{Training Progress}\label{sec:Training Progress}

The entire training process consisted of incomplete supervised learning and pseudo-label-assisted learning. The algorithm is detailed in Algorithm~\ref{alg:algorithm1}. First, we trained the model with initial ground-truth labels for 100 epochs. Then, the initial pseudo labels were generated by the trained model, and we started pseudo-label-assisted learning. When the model converged on the ground truth, we updated the pseudo-labels. This was repeated for 100 epochs.

\begin{figure*}[hb]
\begin{center}
       \subfigure[]{
                \begin{minipage}{.4\linewidth}
                \centering
                \includegraphics[width=1.0\columnwidth]{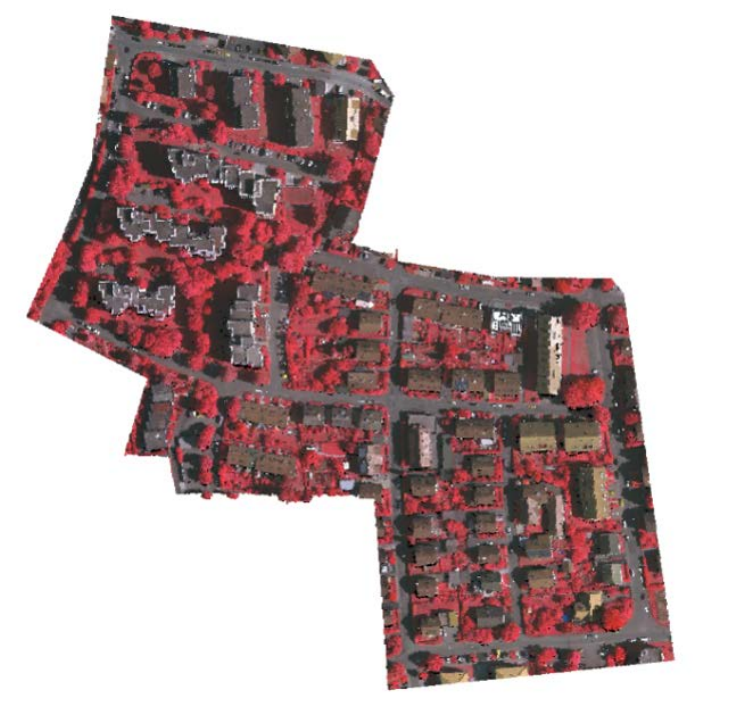}
                \end{minipage}
                }
      \subfigure[]{
                \begin{minipage}{.4\linewidth}
                \centering
                \includegraphics[width=1.0\columnwidth]{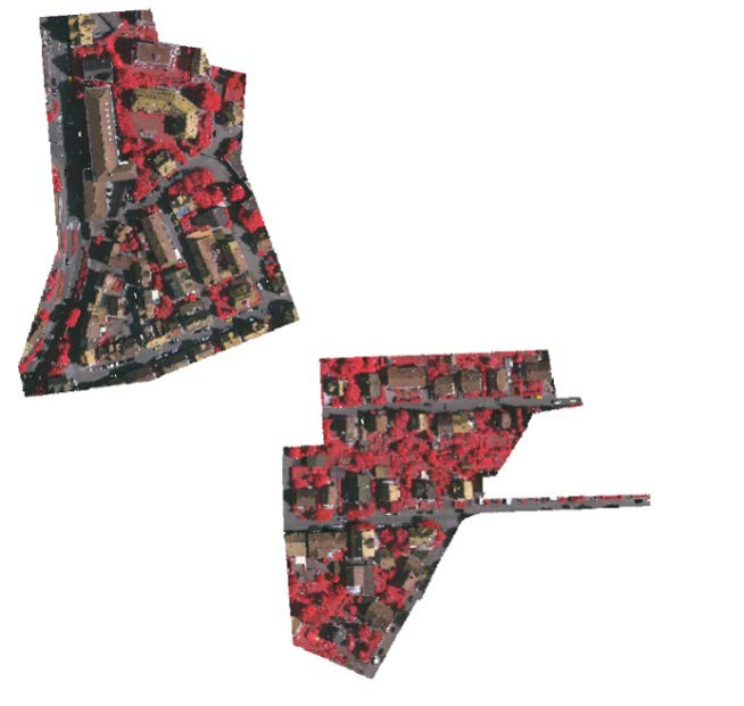}
                \end{minipage}
                } 
      \subfigure[]{
                \begin{minipage}{.4\linewidth}
                \centering
                \includegraphics[width=1.0\columnwidth]{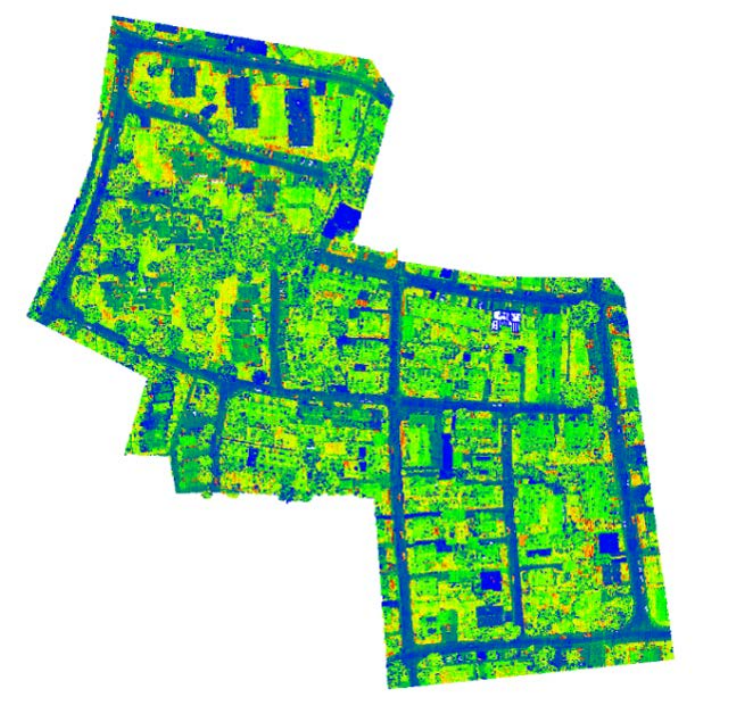}
                \end{minipage}
                }
      \subfigure[]{
                \begin{minipage}{.4\linewidth}
                \centering
                \includegraphics[width=1.0\columnwidth]{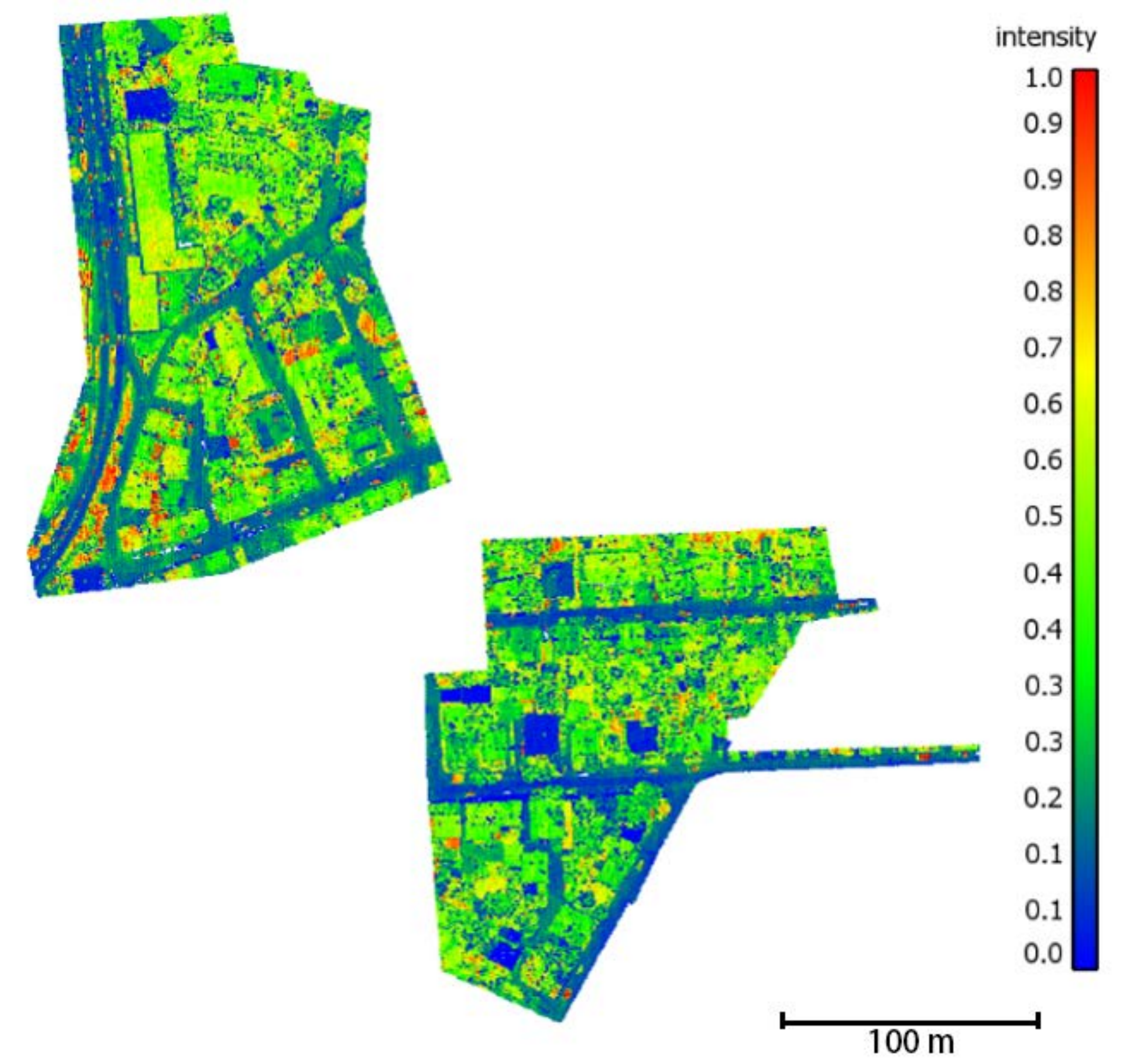}
                \end{minipage}
                }           
	\caption{(a) and (b) represent the color maps (IR, R, G) of training and testing datasets, respectively, whereas (c) and (d) represent the normalized intensity maps of the two sets.}
\label{fig:figure3}
\end{center}
\end{figure*}

\section{EXPERIMENT}\label{sec:EXPERIMENT}

\subsection{Dataset}\label{sec:Dataset}

The ISPRS benchmark dataset~\citep{rottensteiner2014results} was used in this study, as illustrated in Fig.~\ref{fig:figure3}. The dataset contains ALS data obtained from the Leica ALS50 system and the corresponding remote sensing image for extracting color information. ALS data and remote sensing images were obtained from the Stuttgart region of Germany. The point density of the data was between 4 and 7 points per $m^2$. The corresponding remote sensing image covered the entire area, and the ground sampling distance was 8 cm. There were 9 categories in the dataset, including powerlines, low vegetation, impervious surfaces, cars, fences, building roofs, building facades, shrubs, and trees. The data were divided into two sets for training and testing, and the number of points for the training and testing sets was 753,859 and 411,721, respectively. The number of point clouds of different types considerably varied and were mainly distributed in the following four categories: low vegetation, impervious surfaces, building roofs, and trees. This data is listed in Table~\ref{Tab:table1}. 

The dataset contained multiple scan data, and the distance between some points in the overlap area was very small. To remove redundant point clouds in overlapping ALS strips and maintain the even point cloud density, we set the subsampling grid size to d = 0.4 m and assigned labels to deleted points according to the nearest neighbor point in testing. After subsampling, the number of training sets was 401892, and weak labels were selected from the subsampled data. We followed three weak-label selection criteria:

\begin{itemize}
\item[$\bullet$] Weak labels were randomly selected from the training data;
\item[$\bullet$] The number of weak labels in each category did not exceed $10\%$ of the number of that category;
\item[$\bullet$] Less weak-label cases were included in the more weakly labeled cases. For instance, the selected labels in 15 weak-label settings were fully included in the 30 weak-label settings.
\end{itemize}

\begin{table}[t]
\centering
\small
\caption{Class distribution of ISPRS 3D dataset}
\begin{tabular}{ccc}
\toprule[1pt]
Class               & Training set & Testing set \\ \hline
Powerline           & 546          & 600         \\
Low vegetation      & 180,850      & 98,690      \\
Impervious surfaces & 193,723      & 101,986     \\
Car                 & 4,614        & 3,708       \\
Fence/hedge         & 12,070       & 7,422       \\
Roof                & 152,045      & 109,048     \\
Facade              & 27,250       & 11,224      \\
Shrub               & 47,605       & 24,818      \\
Tree                & 135,173      & 54,226      \\ \hline
Total               & 753,876      & 411,722     \\ \bottomrule[1pt]
\end{tabular}
\label{Tab:table1}
\end{table}

Four parameter settings, that is, 15, 30, 60, and 100 indicating 135, 270, 512, and 832 initially labeled points, respectively were examined for weak label numbers. The percentage of 100 weak-label settings to total points was approximately 0.2$\%$.

\begin{figure*}[b]
\captionsetup{justification=centering}
\begin{center}
		\includegraphics{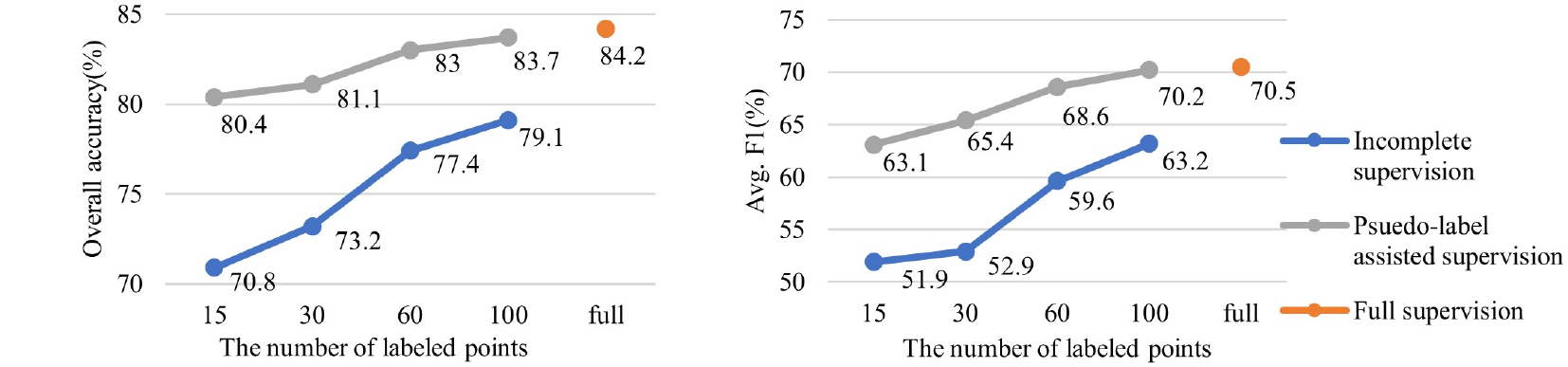}
	\caption{The impact of the number of labeled points on OA and Avg. F1.}
\label{fig:figure4}
\end{center}
\end{figure*}

\begin{table*}[b]
\centering
\small
\caption{Comparison of full supervision and weak supervision}
\begin{tabular}{clcccccccccccc}
    \toprule[1pt]
    \multicolumn{2}{c}{\multirow{2}{*}{Setting}} & \multirow{2}{*}{Method} & \multicolumn{9}{c}{F1 Score}  & \multirow{2}{*}{Avg. F1} & \multirow{2}{*}{OA} \\ \cline{4-12}
    \multicolumn{2}{c}{} & & Power & \begin{tabular}[c]{@{}c@{}}Low \\ veg.\end{tabular} & \begin{tabular}[c]{@{}c@{}}Imp.\\ Surf.\end{tabular} & Car  & Fence & Roof & Facade & Shrub & Tree & & \\ \hline 
    \multicolumn{2}{c}{\multirow{3}{*}{\rotatebox{90}{Full Sup.}}}   
                             & NANJ2  & 62.0  & \textbf{88.8}  & 91.2  & 66.7 & 40.7  & 93.6 & 42.6   & \textbf{55.9}  & 82.6 & 69.3  & \textbf{85.2}  \\
    \multicolumn{2}{c}{}                                                 
                             & WhuY4 & 42.5  & 82.7 & \textbf{91.4}  & 74.7 & \textbf{53.7}  & 94.3 & 53.1   & 47.9  & \textbf{82.8} & 69.2 & 84.9  \\
    \multicolumn{2}{c}{}                                                
                             & KPConv & \textbf{68.7}  & 81.6  & 91.0 & \textbf{75.0} & 31.5  & \textbf{95.0} & \textbf{64.2}   & 44.8  & 82.2 & \textbf{70.5}  & 84.2  \\ \hline
    \multirow{8}{*}{\rotatebox{90}{Weak Sup.}} 
    & \multicolumn{1}{c}{\multirow{2}{*}{15}}
                             & Baseline  & 20.9  & 67.4 & 82.0 & 40.7 & 21.5  & 85.4 & 42.0   & \textbf{37.6}  & 69.6 & 51.9 & 70.8 \\
    & \multicolumn{1}{c}{}                    
                             & PL  &\textbf{68.7}  & \textbf{78.4}  & \textbf{89.1} & \textbf{60.9} & \textbf{25.6}  & \textbf{94.2} & \textbf{54.6}   & 16.0  & \textbf{80.7} & \textbf{63.1}  & \textbf{80.4} \\ \cline{2-14} 
    & \multicolumn{1}{c}{\multirow{2}{*}{30}}
                             & Baseline & 20.8  & 74.7 & 87.9 & 37.1 & 19.8  & 86.1 & 48.7   & \textbf{33.8}  & 67.3 & 52.9 & 73.4 \\
    & \multicolumn{1}{c}{}
                             & PL & \textbf{65.4}  & \textbf{78.2}  & \textbf{89.3}  & \textbf{65.5} & \textbf{27.0}  & \textbf{94.4} & \textbf{58.3}   & 31.5  & \textbf{79.2} & \textbf{65.4} & \textbf{81.1} \\ \cline{2-14} 
    & \multirow{2}{*}{60} 
                             & Baseline & 34.3  & 76.1 & \textbf{90.2}  & 55.8 & 28.8  & 89.5 & 52.2   & 37.9  & 71.8 & 59.6  & 77.4  \\
    &                                        
                             & PL & \textbf{64.0}  & \textbf{79.9} & 89.9 & \textbf{69.6} & \textbf{31.0}  & \textbf{94.6} & \textbf{60.1}   & \textbf{46.6}  & \textbf{81.3} & \textbf{68.6} & \textbf{83.0} \\ \cline{2-14} 
    & \multirow{2}{*}{100} 
                             & Baseline & 45.9 & 75.3 & 89.9 & 58.8 & 32.3  & 93.2 & 60.6   & 36.1  & 76.3 & 63.2  & 79.1 \\
    &                                         
                             & PL & \textbf{68.9}  & \textbf{81.0} & \textbf{90.7} & \textbf{68.0} & \textbf{37.4}  & \textbf{94.7} & \textbf{64.0}   & \textbf{45.8}  & \textbf{81.5} & \textbf{70.2} & \textbf{83.7} \\ \bottomrule[1pt]
\label{Tab:table2}
\end{tabular}
\end{table*}

\subsection{Implementation}\label{sec:Implementation}

During the training, we sliced blocks with a radius of 30 m, and each time the selected central point was the one with the minimum number of times being trained. The batch size and steps models were implemented in the framework of Tensorflow and trained on a GeForce GTX 1080Ti 11 GB GPU. The entire training process lasted approximately 3 h.

\subsection{Evaluation}\label{sec:Evaluation}

Following the evaluation criteria of the ISPRS benchmarks, we used the overall accuracy (OA) and F1 score to evaluate the performance of our method. OA is the percentage of predictions correctly classified. The F1 score is the harmonic mean of the precision and recall:
\begin{equation}\label{equ:6}
\begin{aligned}
& precision = \frac{{tp}}{{tp+ fp}},\\
& recall = \frac{{tp}}{{tp + fn}},\\
& F1 = 2 \times \frac{{precision \times recall}}{{precision + recall}},\\
\end{aligned}
\end{equation}
where $tp$, $fp$, and $fn$ are true positives, false positives, and false negatives, respectively.

\begin{figure*}[t]
\begin{center}
       \subfigure[]{
                \begin{minipage}{0.485\linewidth}
                \centering
                \includegraphics[width=1.0\columnwidth]{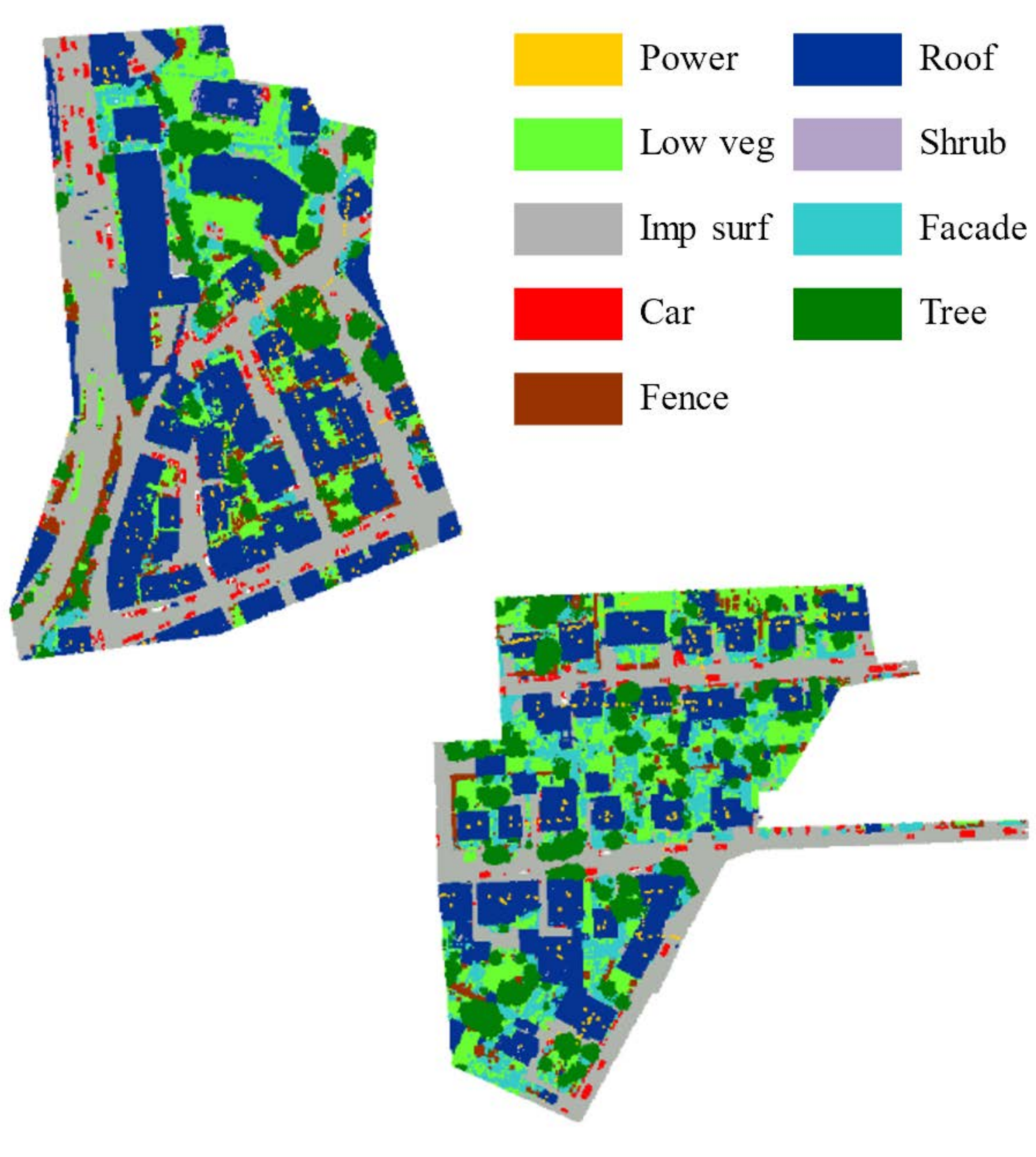}
                \end{minipage}
                }
      \subfigure[]{
                \begin{minipage}{0.485\linewidth}
                \centering
                \includegraphics[width=1.0\columnwidth]{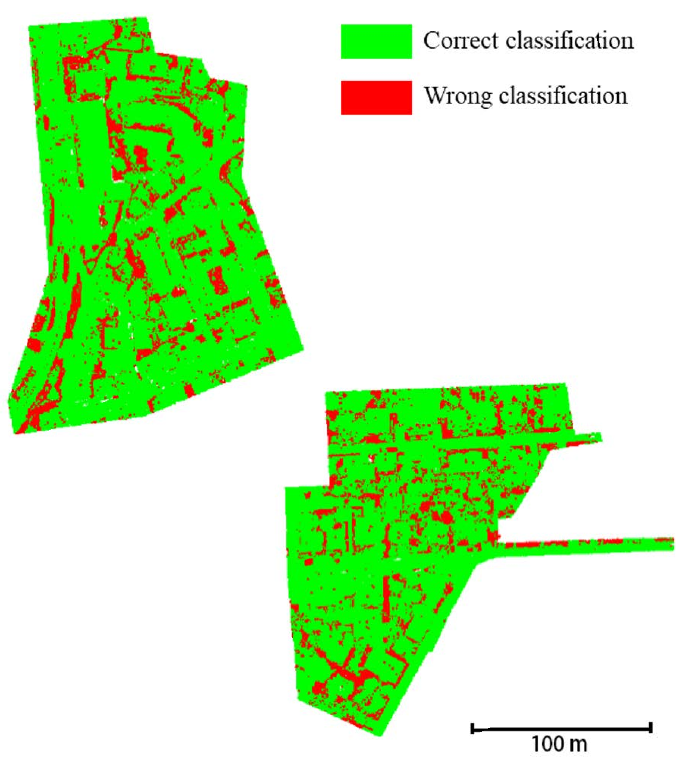}
                \end{minipage}
                }               
	\caption{(a) and (b) represent the classification result and the error map of the proposed method in 100 weak-label setting, respectively.}
\label{fig:figure5}
\end{center}
\end{figure*}

\subsection{Results and analysis}\label{sec:Results and analysis}

We first trained the model under weak-label settings, and pseudo-label-assisted learning was proposed with trained model parameters. For a better comparison, a result was obtained by the KPConv network under full supervision. Figure~\ref{fig:figure4} shows the OA and average F1 score (avg. F1) under the different conditions. OA and Avg. F1 were significantly improved after pseudo-label-assisted learning. The improvement was most evident in 15 weak-label settings, and the OA improved by nearly 10$\%$, from 70.8$\%$ to 80.4$\%$, as well as the Avg. F1, from 51.9$\%$ to 63.1$\%$. In the 100 weak-label setting, the numbers for OA and Avg. F1 were 84.2$\%$ and 70.5$\%$, slightly below those under full supervision. With pseudo-label-assisted learning, we achieved a result comparable to that from full supervision by using approximately 0.2$\%$ of full labels in the training set only. The classification result and error map using 100 weak-label settings are shown in Fig.~\ref{fig:figure5}.

Two methods with leading results were also compared in this work, both using fully supervised learning during training, and the results are presented in Table~\ref{Tab:table2}. NANJ2~\citep{zhao2018classifying} and WhuY4~\citep{yang2018segmentation} had better performance regarding OA, reaching 85.2$\%$ and 84.9$\%$, respectively, whereas KPConv achieved a higher Avg. F1. This result illustrates the effectiveness of KPConv in point cloud semantic segmentation, and it also demonstrates that a competitive result compared to fully supervised methods was achieved with very limited and sparse ground-truth labels. A detailed comparison of the results before and after pseudo-label-assisted learning is presented in Table 2. The F1 score of almost every category was considerably improved under all weak-label settings, except for the class of shrub, in which the number of weak labels was very small, namely 15 weak-label settings. This is a limitation of pseudo-label methods, and leads to considerable errors in the pseudo labels generated from a model with poor performance. Because these false predictions are treated as ground-truth labels for training, when the error rate in certain categories is too high, false pseudo-labels dominate during subsequent training iterations and contribute to a side effect in the model performance with respect to these categories, which provides overconfident predictions~\citep{lokhande2020generating}.

\begin{figure}[t]
\begin{center}
		\includegraphics[width=1.0\columnwidth]{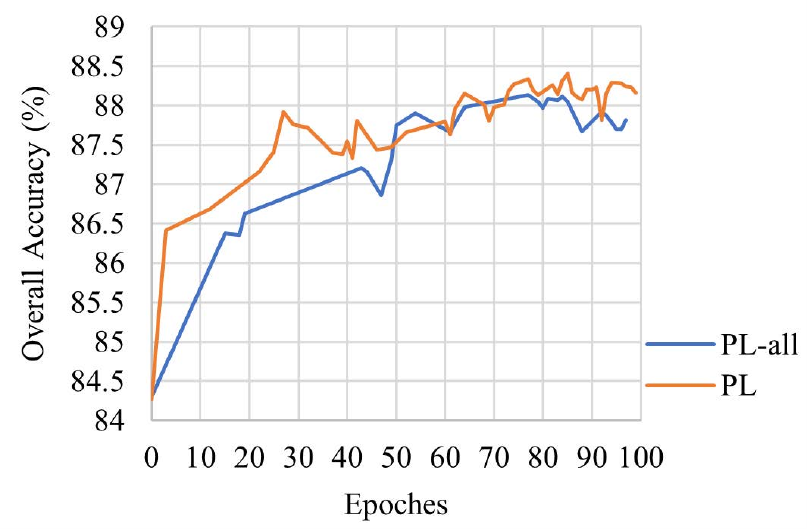}
	\caption{Comparison of overall accuracy of validation on training data in 100 weak-label setting between two pseudo-label update strategies.}
\label{fig:figure6}
\end{center}
\end{figure}

To verify the effectiveness of our pseudo-label update strategy, a comparison experiment was conducted, as shown in Fig.~\ref{fig:figure6}. PL-all represents the strategy in which pseudo labels are updated when the model converges on both weak and pseudo labels, and PL is our method. As pseudo-label-assisted learning is transductive learning in training data, we used the OA of the training data to indicate the training process. We observed that PL achieved a better accuracy in early epochs, which means that the model was trained at a faster speed. In addition, the OA of the PL was higher than that of the PL-all at the end of training. However, because the number of training epochs was fixed at 100, the consumed time was the same for the two strategies. 

\begin{figure*}[h!]
\begin{center}
       \subfigure[]{
                \begin{minipage}{.3\linewidth}
                \centering
                \includegraphics[width=1.0\columnwidth]{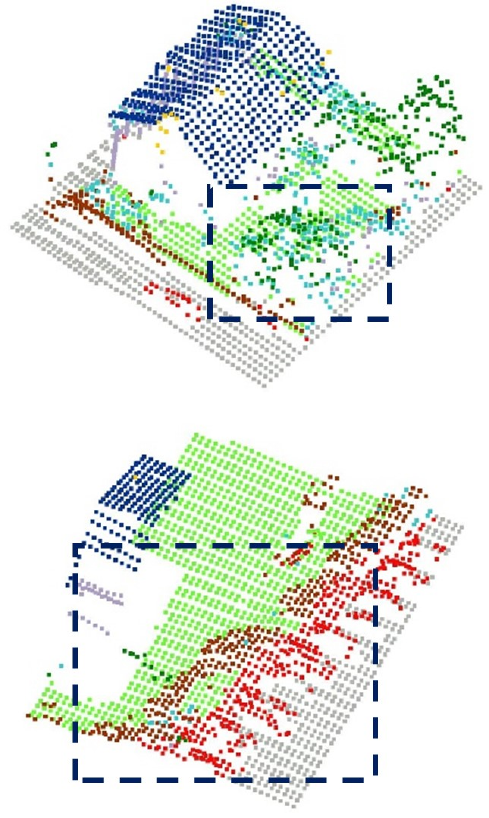}
                \end{minipage}
                }
      \subfigure[]{
                \begin{minipage}{.3\linewidth}
                \centering
                \includegraphics[width=1.0\columnwidth]{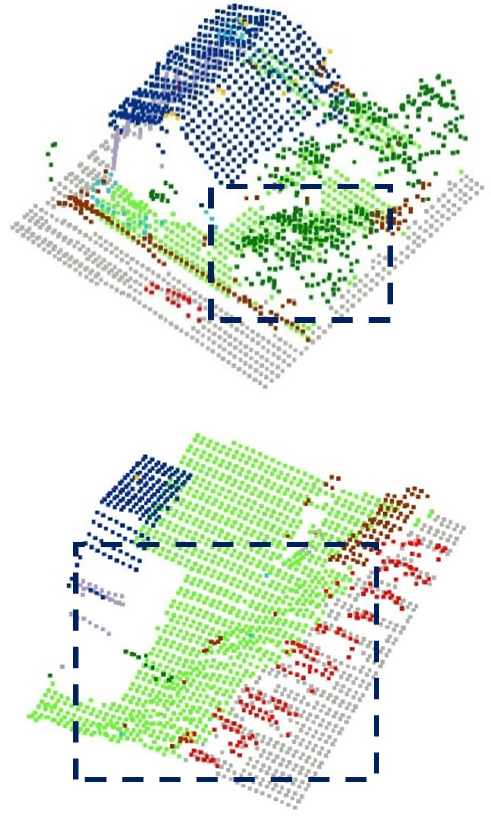}
                \end{minipage}
                } 
      \subfigure[]{
                \begin{minipage}{.3\linewidth}
                \centering
                \includegraphics[width=1.0\columnwidth]{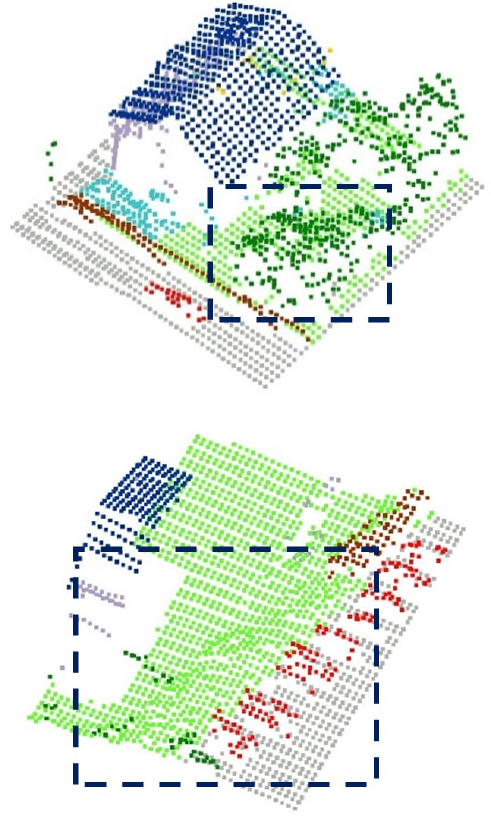}
                \end{minipage}
                }        
	\caption{Details of predictions in training set. (a) and (b) represent the results before and after pseudo-label assisted learning, and (c) is the ground truth.}
\label{fig:figure7}
\end{center}
\end{figure*}

The effect of the pseudo-label method was similar to that of entropy regularization~\citep{grandvalet2005semi} by considering the probability of predictions $P(P\leq1)$ as 1. Therefore, the posteriori probability of predictions will increase through the information transmission of pseudo-labels during training. We found that it is beneficial to label smoothness and reduce noise in the predictions. Figure~\ref{fig:figure7} presents a comparison of the details in the training set before and after pseudo-label-assisted learning. Within the black boxes, the misclassified points of tree and low vegetation were corrected, which indicates that the accuracy of pseudo labels generated from the training set increased along with iterative update steps.

\section{CONCLUSION}\label{sec:CONCLUSION}

In this work, we proposed a pseudo-label-assisted point cloud semantic segmentation method for ALS data in urban areas. KPConv was used as the backbone for the feature extraction. An adaptive threshold was designed to balance the training accuracy and the required amount of pseudo labels. In addition, we used convergence on ground-truth labels as the requirement for updating pseudo-labels, which considerably improved the training efficiency. Experiments indicated that a competitive result was achieved compared to those under the full supervision scheme with only 2$\permil$ of the original abundant labeled points. 

In the future, the limitations of the pseudo-labels mentioned in the experimental analysis are expected to be solved. Because the recursive generation of pseudo labels is completely determined by the probability of the predictions, they are highly correlated with the selected weak labels. Thus, additional constraints are required to control the generation of pseudo-labels. In addition, an elaborated weak-label selection strategy can help achieve better performance under the same number of weak labels, which is a good advance for data annotation.

\section{ACKNOWLEDGEMENTS}\label{sec:ACKNOWLEDGEMENTS}

The dataset was provided by the German Society for Photogrammetry, Remote Sensing, and Geoinformation (DGPF). The work described in this paper was substantially supported by a grant from the Research Grants Council of the Hong Kong Special Administrative Region, China (Project No. PolyU 25211819), and was also supported by a grant from the Hong Kong Polytechnic University (Project No. G-YBZ9).

{
	\begin{spacing}{1.17}
	    \setlength{\bibhang}{0em}
	    \renewcommand\bibsection{\section*{REFERENCES}}
		\normalsize
		\bibliography{main} 

\begin{thebibliography}{xx}

\bibitem[Bearman et al., 2016]{bearman2016s}
Bearman, A., Russakovsky, O., Ferrari, V., Fei-Fei, L., 2016.
 What's the point: Semantic segmentation with point supervision.
 B.~Leibe, J.~Matas, N.~Sebe, M.~Welling (eds), \emph{Computer Vision -- ECCV
  2016}, Springer International Publishing, Cham, 549--565.

\bibitem[Berthelot et al., 2019]{NEURIPS2019_1cd138d0}
Berthelot, D., Carlini, N., Goodfellow, I., Papernot, N., Oliver, A., Raffel,
  C.~A., 2019.
 Mixmatch: A holistic approach to semi-supervised learning.
 H.~Wallach, H.~Larochelle, A.~Beygelzimer, F.~d\textquotesingle Alch\'{e}-Buc,
  E.~Fox, R.~Garnett (eds), \emph{Advances in Neural Information Processing
  Systems}, ~32, Curran Associates, Inc.

\bibitem[Boulch et al., 2018]{boulch2018snapnet}
Boulch, A., Guerry, J., {Le Saux}, B., Audebert, N., 2018.
 SnapNet: 3D point cloud semantic labeling with 2D deep segmentation networks.
 {\em Computers \& Graphics}, 71, 189-198.

\bibitem[Chapelle et al., 2010]{chapelle2009semi}
Chapelle, O., Schlkopf, B., Zien, A., 2010.
 {\em Semi-Supervised Learning}.
 1st edn, The MIT Press.

\bibitem[Dong and Xing, 2018]{dong2018few}
Dong, N., Xing, E.~P., 2018.
 Few-shot semantic segmentation with prototype learning.
 \emph{BMVC},  3(4), 1--13.

\bibitem[Grandvalet et al., 2005]{grandvalet2005semi}
Grandvalet, Y., Bengio, Y. et~al., 2005.
 Semi-supervised learning by entropy minimization.
 \emph{CAP}, 281--296.

\bibitem[Guinard and Landrieu, 2017]{guinard2017weakly}
Guinard, S., Landrieu, L., 2017.
 WEAKLY SUPERVISED SEGMENTATION-AIDED CLASSIFICATION OF URBAN SCENES FROM 3D
  LIDAR POINT CLOUDS.
 {\em The International Archives of the Photogrammetry, Remote Sensing and
  Spatial Information Sciences}, XLII-1/W1, 151--157.
 https://www.int-arch-photogramm-remote-sens-spatial-inf-sci.net/XLII-1-W1/151/2017/.

\bibitem[Hu et al., 2020]{hu2020randla}
Hu, Q., Yang, B., Xie, L., Rosa, S., Guo, Y., Wang, Z., Trigoni, N., Markham,
  A., 2020.
 Randla-net: Efficient semantic segmentation of large-scale point clouds.
 \emph{Proceedings of the IEEE/CVF Conference on Computer Vision and Pattern
  Recognition}, 11108--11117.

\bibitem[Huang et al., 2020]{huang2020deep}
Huang, R., Xu, Y., Hong, D., Yao, W., Ghamisi, P., Stilla, U., 2020.
 Deep point embedding for urban classification using ALS point clouds: A new
  perspective from local to global.
 {\em ISPRS Journal of Photogrammetry and Remote Sensing}, 163, 62--81.

\bibitem[Iscen et al., 2019]{iscen2019label}
Iscen, A., Tolias, G., Avrithis, Y., Chum, O., 2019.
 Label propagation for deep semi-supervised learning.
 \emph{Proceedings of the IEEE/CVF Conference on Computer Vision and Pattern
  Recognition}, 5070--5079.

\bibitem[Kolesnikov and Lampert, 2016]{kolesnikov2016seed}
Kolesnikov, A., Lampert, C.~H., 2016.
 Seed, expand and constrain: Three principles for weakly-supervised image
  segmentation.
 B.~Leibe, J.~Matas, N.~Sebe, M.~Welling (eds), \emph{Computer Vision -- ECCV
  2016}, Springer International Publishing, Cham, 695--711.

\bibitem[Lee, 2013]{lee2013pseudo}
Lee, D.-H., 2013.
 Pseudo-label: The simple and efficient semi-supervised learning method for
  deep neural networks.
 \emph{Workshop on challenges in representation learning, ICML},  3(2), 1--6.

\bibitem[Lin et al., 2020]{lin2020active}
Lin, Y., Vosselman, G., Cao, Y., Yang, M.~Y., 2020.
 Active and incremental learning for semantic ALS point cloud segmentation.
 {\em ISPRS Journal of Photogrammetry and Remote Sensing}, 169, 73--92.

\bibitem[Lokhande et al., 2020]{lokhande2020generating}
Lokhande, V.~S., Tasneeyapant, S., Venkatesh, A., Ravi, S.~N., Singh, V., 2020.
 Generating accurate pseudo-labels in semi-supervised learning and avoiding
  overconfident predictions via hermite polynomial activations.
 \emph{Proceedings of the IEEE/CVF Conference on Computer Vision and Pattern
  Recognition}, 11435--11443.

\bibitem[Papandreou et al., 2015]{papandreou2015weakly}
Papandreou, G., Chen, L.-C., Murphy, K.~P., Yuille, A.~L., 2015.
 Weakly-and semi-supervised learning of a deep convolutional network for
  semantic image segmentation.
 \emph{Proceedings of the IEEE international conference on computer vision},
  1742--1750.

\bibitem[Polewski et al., 2015]{polewski2015active}
Polewski, P., Yao, W., Heurich, M., Krzystek, P., Stilla, U., 2015.
 Active learning approach to detecting standing dead trees from als point
  clouds combined with aerial infrared imagery.
 \emph{Proceedings of the IEEE Conference on Computer Vision and Pattern
  Recognition Workshops}, 10--18.

\bibitem[{Qi} et al., 2017a]{qi2017pointnet}
{Qi}, C.~R., {Su}, H., {Kaichun}, M., {Guibas}, L.~J., 2017a.
 Pointnet: Deep learning on point sets for 3d classification and segmentation.
 \emph{2017 IEEE Conference on Computer Vision and Pattern Recognition (CVPR)},
  77--85.

\bibitem[Qi et al., 2017b]{qi2017pointnet++}
Qi, C.~R., Yi, L., Su, H., Guibas, L.~J., 2017b.
 Pointnet++ deep hierarchical feature learning on point sets in a metric space.
 \emph{Proceedings of the 31st International Conference on Neural Information
  Processing Systems}, 5105--5114.

\bibitem[Rottensteiner et al., 2014]{rottensteiner2014results}
Rottensteiner, F., Sohn, G., Gerke, M., Wegner, J.~D., Breitkopf, U., Jung, J.,
  2014.
 Results of the ISPRS benchmark on urban object detection and 3D building
  reconstruction.
 {\em ISPRS journal of Photogrammetry and Remote Sensing}, 93, 256--271.

\bibitem[Steinsiek et al., 2017]{steinsiek2017semantische}
Steinsiek, M., Polewski, P., Yao, W., Krzystek, P., 2017.
 Semantische analyse von als-und mls-daten in urbanen gebieten mittels
  conditional random fields.
 \emph{Wissenschaftlich-Technische Jahrestagung der DGPF}, ~37, 521--531.

\bibitem[{Tchapmi} et al., 2017]{tchapmi2017segcloud}
{Tchapmi}, L., {Choy}, C., {Armeni}, I., {Gwak}, J., {Savarese}, S., 2017.
 Segcloud: Semantic segmentation of 3d point clouds.
 \emph{2017 International Conference on 3D Vision (3DV)}, 537--547.

\bibitem[Thomas et al., 2019]{thomas2019kpconv}
Thomas, H., Qi, C.~R., Deschaud, J.-E., Marcotegui, B., Goulette, F., Guibas,
  L.~J., 2019.
 Kpconv: Flexible and deformable convolution for point clouds.
 \emph{Proceedings of the IEEE/CVF International Conference on Computer
  Vision}, 6411--6420.

\bibitem[Wang et al., 2019]{wang2019dynamic}
Wang, Y., Sun, Y., Liu, Z., Sarma, S.~E., Bronstein, M.~M., Solomon, J.~M.,
  2019.
 Dynamic Graph CNN for Learning on Point Clouds.
 {\em ACM Trans. Graph.}, 38(5).
 https://doi.org/10.1145/3326362.

\bibitem[Wei et al., 2020]{wei2020multi}
Wei, J., Lin, G., Yap, K.-H., Hung, T.-Y., Xie, L., 2020.
 Multi-path region mining for weakly supervised 3d semantic segmentation on
  point clouds.
 \emph{Proceedings of the IEEE/CVF Conference on Computer Vision and Pattern
  Recognition}, 4384--4393.

\bibitem[Xu and Lee, 2020]{xu2020weakly}
Xu, X., Lee, G.~H., 2020.
 Weakly supervised semantic point cloud segmentation: Towards 10x fewer labels.
 \emph{Proceedings of the IEEE/CVF Conference on Computer Vision and Pattern
  Recognition}, 13706--13715.

\bibitem[Yang et al., 2018]{yang2018segmentation}
Yang, Z., Tan, B., Pei, H., Jiang, W., 2018.
 Segmentation and multi-scale convolutional neural network-based classification
  of airborne laser scanner data.
 {\em Sensors}, 18(10), 3347.

\bibitem[Yao et al., 2020]{yao2020pseudo}
Yao, Y., Xu, K., Murasaki, K., Ando, S., Sagata, A., 2020.
 Pseudo-labelling-aided semantic segmentation on sparsely annotated 3D point
  clouds.
 {\em IPSJ Transactions on Computer Vision and Applications}, 12(1), 1--13.

\bibitem[Zhao et al., 2018]{zhao2018classifying}
Zhao, R., Pang, M., Wang, J., 2018.
 Classifying airborne LiDAR point clouds via deep features learned by a
  multi-scale convolutional neural network.
 {\em International Journal of Geographical Information Science}, 32(5),
  960--979.

\bibitem[Zhou, 2017]{zhou2018brief}
Zhou, Z.-H., 2017.
 {A brief introduction to weakly supervised learning}.
 {\em National Science Review}, 5(1), 44-53.
 https://doi.org/10.1093/nsr/nwx106.

\bibitem[Zhu, 2005]{zhu2005semi}
Zhu, X.~J., 2005.
 Semi-supervised learning literature survey.

\bibitem[Zou et al., 2021]{zou2020pseudoseg}
Zou, Y., Zhang, Z., Zhang, H., Li, C.-L., Bian, X., Huang, J.-B., Pfister, T.,
  2021.
 Pseudoseg: Designing pseudo labels for semantic segmentation.

\end{thebibliography}
	\end{spacing}
}

\end{document}